# Classifier Transfer with Data Selection Strategies for Online Support Vector Machine Classification with Class Imbalance

**Mario Michael Krell**[1]**, Nils Wilshusen**[1]**, Anett Seeland**[2] **and Su Kyoung Kim**[2]

[1] Robotics Research Group, University of Bremen, Robert-Hooke-Str. 1, Bremen, Germany
[2] Robotics Innovation Center, German Research Center for Artificial Intelligence GmbH, Bremen, Germany

E-mail: `krell@uni-bremen.de`

**Abstract.** *Objective:* Classifier transfers usually come with dataset shifts. To overcome dataset shifts, online strategies have to be applied. For practical applications, limitations in the computational resources for the adaptation of batch learning algorithms, like the support vector machine (SVM), have to be considered. *Approach:* We review and compare several strategies for online learning with SVMs. We focus on data selection strategies which limit the size of the stored training data, by different inclusion, exclusion, and further dataset manipulation criteria. First, we provide a comparison of the strategies on several synthetic datasets with different data shifts. Secondly, we analyze the methods on different transfer settings with EEG data. When processing real world data, class imbalance occurs quite often, e.g., in oddball experiments. This may also result from the data selection strategy itself. We analyze this effect, by evaluating two new balancing criteria. *Main Results:* For different data shifts, different criteria are appropriate. For the synthetic data, adding all samples to the pool of considered samples performs often significantly worse than other criteria. Especially, adding only misclassified samples performed astoundingly well. Here, balancing criteria were very important when the other criteria were not well chosen. For the transfer setups, the results show that the best strategy depends on the intensity of the drift during the transfer. Adding all and removing the oldest samples results in the best performance, whereas for smaller drifts, it can be sufficient to only add potential new support vectors of the SVM which reduces processing resources. *Significance:* For brain-computer interfaces based on electroencephalographic (EEG) models, trained on data from a calibration session, a previous recording session, or even from a recording session with one or several other subjects, are used. This transfer of the learned model usually decreases the performance and can therefore benefit from online learning which adapts the classifier like the established SVM. We show that by using the right combination of data selection criteria, it is possible to adapt the classifier and largely increase the performance. Furthermore, in some cases it is possible to speed up the processing and save computational by updating with a subset of special samples and keeping a small subset of samples for training the classifier.

Submitted to: *J. Neural Eng.*

# 1. INTRODUCTION

## 1.1. Motivation

The support vector machine (SVM) has become a well known classification algorithm due to its good performance (Cristianini & Shawe-Taylor 2000, Müller et al. 2001, Schölkopf & Smola 2002, Vapnik 2000). The SVM is a static batch learning algorithm, i.e., it uses all training data to build a model and does not change when new data is processed. Despite having the advantage of being a powerful and reliable classification algorithm, the SVM can run into problems when dataset shifts (Quionero-Candela et al. 2009) occur due to its static nature. The issues is amplified when the SVM should runs on systems with limited resources, e.g., on a mobile device, because a complete retraining is probably not possible when new training data is acquired.

In the context of SVM learning applied to encephalographic data (EEG), as used for brain-computer interfaces (BCIs) (Blankertz et al. 2011, Zander & Kothe 2011, Kirchner et al. 2013, Wöhrle et al. 2015), dataset shifts are a major issue. The source of the problem lies in the fact that the observed EEG pattern changes over time, e.g., due to inherent conductivity fluctuations, sensor displacement, or subject tiredness. The issue of dataset shifts also occurs in other applications like robotics. A classic example of a dataset shift would be that of a computer vision classifier being trained during daylight and then used to classify image data taken during the night. Temperature fluctuations, wear and tear inside the robotic mechanics can lead to a different behavior in certain robotic systems. The behavioral change then impacts measurement results, which also leads to dataset shifts. Dataset shifts can occur in a continuous way but also in a more abrupt way. The research area of transfer learning analyses the dataset shift, that occurs, when models are transferred from a (usually controlled) training setting to a real application or just a related but different setting.

## 1.2. Transfer learning for BCIs

Transfer learning is a relevant issue in BCI applications, which uses historic data from different sessions, subjects, tasks, and even different event-related potential (ERP) types to reduce calibration time.

Recent studies reported different transfer scenarios between different tasks and between different types of error-related potentials (ErrP) (Iturrate et al. 2012, Iturrate et al. 2013, Kim & Kirchner 2013, Kim & Kirchner 2016, Spüler & Niethammer 2015). The basis of these transfers is to either use similar ERP shapes or correct differences, e.g., latency correction or feature selection. Hence the dataset shift shall be reduced.

Several other studies investigated the session-to-session transfer of spatial filter and classifier and reported that calibration time, i.e, time for recording an additional training dataset, was clearly reduced with a minor loss in classification performance compared to session-specific calibration, e.g., (Krauledat et al. 2008). But the transfer of spatial filters between subjects is problematic. A common approach is to either combine classifiers, e.g., with an ensemble approach (Fazli et al. 2009, Metzen et al. 2011), or to improve the spatial filter (Blankertz et al. 2007, Lotte & Guan 2011, Krell, Wöhrle & Seeland 2015, Wöhrle et al. 2015).

Despite many positive examples of transfer learning, all of them still result in a dataset shift and consequently in a decrease of classification performance.

### 1.3. Online learning approaches to handle dataset shifts

No matter what is the reason for these shifts in the data, it is often possible to adapt the classifier for new incoming data to increase again the performance. A straightforward approach to handle the dataset shift would be to integrate a mechanism, that labels the incoming data, and afterwards use the labeled data to retrain the SVM. Such a retraining approach is used by (Steinwart et al. 2009), whereby the optimization problem is given a warm start, i.e., the old solution is used to generate a good starting point for the iterative solution algorithm. Note, that there are lots of other algorithms which implement incremental SVMs more efficiently but they are not subject of this paper (Laskov et al. 2006, Liang & Li 2009). However, there are computational drawbacks that come with incremental SVMs, since updates only add but not delete samples. This leads to an increase in both processing time (especially when kernels are used), as well as in memory consumption which is the major problem. Taking into account that for applications like for example BCIs and robotics, mobile devices are often used for processing (Wöhrle et al. 2013, Wöhrle et al. 2014), here the lack of computing resources conflicts with a high computational cost. Especially, the online classifier adaptation scheme needs to be faster than the time interval between two incoming samples.

Apart from incremental SVMs, there are numerous other algorithms for efficient online learning like perceptrons (Rosenblatt 1958, Freund & Schapire 1999), linear discriminant analysis (McFarland et al. 2011, Schlögl et al. 2010, Shenoy et al. 2006), stochastic gradient descent based prediction (Zhang 2004), and naïve bayes (Russell & Norvig 2003). The use of online classification- and filter-adaptation has been analyzed in (Wöhrle et al. 2015). In this work, we focus on SVMs (including the online passive-aggressive algorithm as described in Section 2.2) due to its power in generalization especially on high-dimensional data and because it is a common classifier for EEG-data processing.

We focus on a small subgroup of online learning algorithms which limit the size of the training dataset, such that powerful generalization capabilities of SVM are not lost. Even though we will look at SVMs, the scheme can be transferred to any batch learning algorithm that can be speeded up with a warm start. In the context of data shifts, class imbalance is a major issue which has not yet been sufficiently addressed for online learning paradigms (Hoens et al. 2012). Hence, we additionally suggest approaches which tackle this specific problem, namely the case in which:

(i) The class ratio is *ignored*.

(ii) After the initialization, the class ratio is kept *fixed*.

(iii) A *balanced* ratio is strived for throughout the learning process.

## 1.4. Structure

In Section 2, we provide a systematic overview over the different strategies for restricting the size of the training set. We provide a comparison of numerous strategies by testing them, first, on synthetic datasets with different shifts and, second, on a more complex classification task with EEG data in Section 3. This is followed by comparisons on different transfer setups in Section 4. Finally, we conclude in Section 5.

A preliminary version of this paper has been presented at the Neurotechnix conference (Krell, Wilshusen, Ignat & Kim 2015). This version is more elaborate and largely extends the evaluation and presents completely new findings especially on EEG data.

## 2. REVIEW OF TRAINING DATA SELECTION STRATEGIES (FOR THE SVM)

This section introduces the SVM and the different methods for manipulating its training data for online learning. The data handling methods for the SVM can be divided into criteria for adding samples and removing samples, as well as variants which influence both.

### 2.1. Support vector machines (SVMs)

Let $w \in \mathbb{R}^m$ be the classification vector which, together with the offset $b$, defines the linear classification function $f(x) = \langle w, x \rangle + b$. The main part of the SVM concept is the maximum margin which results in the regularization term $\left( \frac{1}{2} \langle w, w \rangle \right)$ (Vapnik 2000). Given the training data $D = \left\{ (x_j, y_j) \in \{-1, +1\} \times \mathbb{R}^m \middle| j \in \{1, \dots, n\} \right\}$, the objective is to maximize the distance between two parallel hyperplanes ($f \equiv 1$ and $f \equiv -1$) which separate samples $x_i$ with positive labels ($y_i = +1$) from samples with negative labels ($y_i = -1$). The second part is the soft margin which allows for some misclassification by means of the loss term, $\sum t_j$. Both parts are weighted with a cost hyperparameter $C$. This is also depicted in Figure 1.

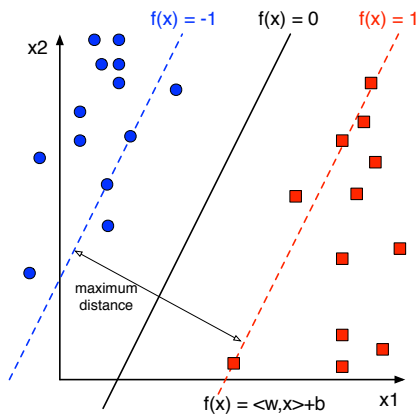

Figure 1: Scheme of the support vector machine (SVM). The red dots are training samples with $y = +1$ and the blue dots with $y = -1$. Displayed are the decision hyperplane in the middle and the two separating hyperplanes with maximum distance considering that some error in the classification is allowed (samples between the blue and the red hyperplane).

Since the data is usually normalized (centered around the origin), the decision hyperplane ($f(x) \equiv 0$) can be expected to be close to the origin with only a small offset, $b$. Hence, the solution is often simplified by minimizing $\frac{1}{2}b^2$ (Hsieh et al. 2008, Mangasarian & Musicant 1998) additionally to $\frac{1}{2}\|w\|$. This simplification can be also obtained by omitting the offset and artificially regaining it by adding an extra component with a 1 to the input feature vectors $x_j$ (Steinwart et al. 2009). The resulting model reads:

$$
\begin{aligned}
\min_{w,b,t} \quad & \tfrac{1}{2}\|w\|_2^2 + \tfrac{1}{2}b^2 \quad + C\sum t_j \\
\text{s.t.} \quad & y_j(\langle w, x_j\rangle + b) \geq 1 - t_j \quad, \forall j : 1 \leq j \leq n, \\
& t_j \geq 0 \qquad\quad, \forall j : 1 \leq j \leq n.
\end{aligned}
\tag{1}
$$

The final label is assigned by using the signum function on $f$. In order to simplify the constraints, and to ease the implementation of the SVM, the dual optimization is often used:

$$
\min_{C \geq \alpha_j \geq 0} \frac{1}{2} \sum_{i,j} \alpha_i \alpha_j y_i y_j (k(x_i, x_j) + 1) - \sum_j \alpha_j.
\tag{2}
$$

The scalar product is replaced by a symmetric, positive, semi-definite kernel function $k$, which is the third part of the SVM concept and allows for nonlinear separation with the decision function

$$
f(x) = \sum_{i=1}^{n} \alpha_i y_i (k(x, x_i) + 1).
\tag{3}
$$

A special property of the SVM is its sparsity in the sample domain. In other words, there is usually a low number of samples which lie exactly on their separating hyperplanes (with $C \geq \alpha_i > 0$), or on the wrong side of their corresponding hyperplane ($\alpha_i = C$, $y_i(\langle w, x_i\rangle + b) < 1$). These samples are the only factors influencing the definition of $f$. All samples with $\alpha > 0$ are called *support vectors*.

### 2.2. Single iteration concept

For the mathematical program of SVM, it can be seen that all training data is required for building the model, even though a reduced number is only needed for defining the classification function. When a new sample is added to the training set, the old sample weights can be reused and updated which is called warm start (Steinwart et al. 2009). The fact that all the data can be relevant is not changed by a new incoming sample. In the case of a linear kernel, one approach for online learning is to calculate only the optimal $\alpha_{n+1}$ if a new sample $x_{n+1}$ comes in and leave the other weights fixed. This update is directly integrated into the calculation of $w$ and $b$, and then the information can be removed from memory, since it is not required anymore. The direct update formula can only be obtained due to the aforementioned offset treatment and because the respective optimization problem can be solved within one step. Otherwise, a different sample weight would have to be reduced and further iterations would be required. The resulting algorithm is also called online passive-aggressive algorithm (PA) (Crammer et al. 2006, Krell et al. 2014). The general principle of offset treatment combined with a single update step to obtain a simple online learning algorithm with a linear classification function is called single iteration concept and it is described in detail in (Krell 2015). The advantage of this approach is, that only the classification vector and the offset are kept in memory. But the approach is not capable of forgetting the influence of the previous samples and it cannot be used together with kernel functions.

Another possibility is to add the new sample, remove another sample, if too many samples are kept in memory, and then retrain the classifier with a limited number of iterations. The numerous existing criteria for this manipulation strategies are introduced in the following sections.

### 2.3. Inclusion criteria

The most common approach is to *add all* samples to the training dataset (Bordes et al. 2005, Funaya et al. 2009, Gretton & Desobry 2003, Oskoei et al. 2009, Tang et al. 2006, Van Vaerenbergh et al. 2010, Van Vaerenbergh et al. 2006, Yi et al. 2011). If the new sample is already on the correct side of its corresponding hyperplane $(y_{n+1}f(x_{n+1}) > 1)$, the classification function will not be changed with the update. When a sample is on the wrong side of the hyperplane, the classification function will change and samples which previously did not have any influence might become important. To reduce the number of updates, there are approaches which only add the samples with importance to the training data. One of these approaches is to add only *misclassified* samples (Bordes et al. 2005, Dekel et al. 2008, Oskoei et al. 2009).

If the true label is unknown, the unsupervised approach by (Spüler et al. 2012) suggests to use the improved Platt's probability fit (Lin et al. 2007) to obtain a probability score from SVM. If the probability exceeds a certain predefined threshold (0.8 in (Spüler et al. 2012)) the label is assumed to be true and the label, together with the sample, are added to the training set. This approach is computationally expensive, since the probability fit has to be recalculated after each classifier update. Furthermore, it its quite inaccurate, because it is calculated on the training data (Lin et al. 2007). Note that, in this approach, samples within the margin are excluded from the update, and that this approach does not consider the maximum margin concept of theSVM.

In contrast to the previous approach, if the true label is known, samples *within the margin* are especially relevant for an update of the SVM (Bordes et al. 2005, Oskoei et al. 2009); PA does the same intrinsically. Data outside the margin gets assigned a weight of zero $(\alpha_{n+1} = 0)$, and so the sample will not be integrated into the classification vector *w*. This method is also closely connected to a variant where all data, which is not a support vector, is removed.

In (Nguyen-Tuong & Peters 2011) a sample is added to the dataset if it is sufficiently linear independent. This concept is generalized to classifiers with kernels. In case of low dimensional data, this approach is not appropriate, while for higher dimensional data, it is computationally expensive. Furthermore, it does not account for the SVM modeling perspective where it is more appropriate to consider the support vectors instead of linear independence. A strong component of the SVM is the margin which is also not considered in the approach. Last but not least it has to be noted, that the SVM also shows good performance when less training data is provided than their is data in the training set. In that application case, the linear independence approach could not reduce any data.

To save resources, a variant for adding samples would be to add a change detection test (CDT), as an additional higher-level layer (Alippi et al. 2014). The CDT detects if there is a change in the data, and then activates the update procedure. When working with datasets with permanent/continuous shifts over time, this approach is not appropriate, since the CDT would always activate the update.

### 2.4. Exclusion criteria

To keep the size of the training set bounded for a fixed batch size, samples have to be removed. One extreme case is the PA which removes the sample directly after adding its influence to the classifier. In other words, the sample itself is discarded, but the classifier remembers its influence. If no additional damping factor in the update formula is used, the influence of the new sample is permanent.

In (Funaya et al. 2009), older samples get a lower weight in the SVM model (exponential decay with a fixed factor). This puts a very large emphasis on new training samples and, at some stage, the weight for the oldest samples is so low, that these samples can be removed. In (Gretton & Desobry 2003), the *oldest* sample is removed for the one-class SVM (Schölkopf et al. 2001). Removing the oldest sample is also closely related to batch updates (Hoens et al. 2012). Here, the classification model remains fixed, while new incoming samples are added to a new training set with a maximum batch size. If this basket is full, all the old data is removed and the model is replaced with a new one, trained on the new training data.

The further away a sample is from the decision boundary, the lower will the respective dual variable be. Consequently, it is reasonable, to remove the farthest sample (Bordes et al. 2005), because it has the lowest impact on the decision function $f$. If the sample $x$ has the function value $|f(x)| > 1$, the respective weight is zero.

In the case of a linear SVM with two strictly separable datasets corresponding to the two classes, the SVM can also be seen as the construction of a separating hyperplane between the convex hulls of the two datasets. So, the *border points* are most relevant for the model, and might become support vectors in future updates. Hence, another criterion for removing data points is to determine the centers of the data and remove all data outside of the two annuli around the centers (Yi et al. 2011). If the number of samples is to high, the weighted distance from the algorithm could be used as a further criterion for removal. Alternatively, we suggest to construct a ring instead of an annulus, samples could be weighted by their distance to the ring and thus, be removed if they are not close enough to the respective ring. Drawbacks of this method are the restriction to linear kernels, the (usually) wrong assumption of circular shapes of the datasets (implied by the use of the annuli), and the additional parameters which are difficult to determine.

Similar to the inclusion criterion, linear independence could also be used for removing the "least linearly independent" samples (Nguyen-Tuong & Peters 2011). This approach suffers from the drawbacks of computational cost and additional hyperparameters.

## 2.5. Further criteria

As already mentioned, support vectors are crucial for the decision function. A commonly used selection criterion is that of *keeping only support vectors* (KSV) (Bordes et al. 2005, Yi et al. 2011) and removing all the other data from the training set.

While in the supervised setting, there might be some label noise from the data sources, in the unsupervised case labels might be assigned completely wrong. For compensation, (Spüler et al. 2012) suggests to *relabel* (REL) every sample with the predicted label from the updated classifier. This approach is also repeatedly used by (Li et al. 2008) for the semi-supervised training of an SVM.

Insofar, the presented methods did not consider the class distribution. The number of samples of one class could be drastically reduced, when the inclusion criterion mostly adds data from one class and/or the exclusion criterion mostly removes data from the other class. Furthermore, when removing samples, it might occur that older data ensured a balance in the class distribution while the incoming data belongs to only one class. Hence, we suggest three different approaches for *data balancing* (BAL).

(i) *Don't handle* the classes differently.

(ii) *Keep the ratio* as it was when first filling the training data basket, i.e., after the initialization always remove a sample from the same class type as it was added in the current update step.

(iii) For us, the most promising approach is to strive for a *balanced ratio* when removing data, by always removing samples from the overrepresented class.

Note that these three criteria can be combined with each other, as well as with all inclusion and exclusion criteria.

## 3. PRE-EVALUATION

This section describes an empirical comparison of a selection of the aforementioned methods on synthetic and EEG data. We start by describing the data generation of synthetic data and data acquisition of EEG data. Afterwards, we present the processing methods and describe the results of the analysis.

## 3.1. Synthetic Data

For the synthetic data, we focus on linearly separable data. The first considered case ("Parallel") is that of a data shift which is parallel to a hyperplane which would be the optimal decision hyperplane. In all the other cases, the optimal decision function changes over time/samples. Five datasets with different shifts are depicted in Figure 2. The data was randomly generated with Gaussian distributed noise and shifting means. Since class distributions are usually unbalanced in reality, we used a ratio of $1:3$ between the two classes, with the underrepresented class labeled as C2.

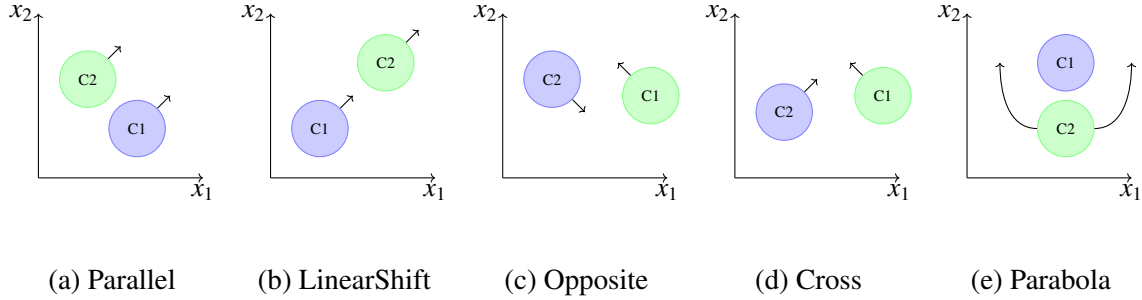

(a) Parallel  (b) LinearShift  (c) Opposite  (d) Cross  (e) Parabola

Figure 2: Synthetic two-dimensional datasets with the classes *C*1 and *C*2 (underrepresented target class) with shifting Gaussian distribution depicted by arrows (see also Section 3.1).

For a three dimensional example, we followed the approach by (Street & Kim 2001) in order to create a dataset with an abrupt concept change every 100 samples, in contrast to the continuous shifts. The data was randomly sampled in a three-dimensional cube. Given a three dimensional sample $(p_1, p_2, p_3)$, a two step procedure is implemented to define the class label. For samples of the first class, initially, $p_1 + p_2 \leq \theta$ has to hold for some given threshold $\theta$ which is randomly changed every 100 samples in the interval $(6, 14)$. Next, 10% class noise is added for both classes. $p_3$ only introduces noise and has no influence on the class.

The six datasets have a total of 10000 samples. Of these, the first 1000 samples are taken for training and hyperparameter optimization while the remaining 9000 samples are used for testing. The same synthetic data was used for all evaluations.

### 3.2. Electroencephalographic data of a P300 scenario

For real data, we used EEG data from P300 oddball paradigm (Courchesne et al. 1977) as described in (Kirchner et al. 2013). In the experimental paradigm, which is intended for use in a real BCI, the subject sees irrelevant information (standard) every second with some jitter. With a probability of $1/6$, important information (target) was displayed, which requires an action from the subject. This task-relevant event leads to a specific pattern in the brain, called P300 (see also Figure 3 for an average curve at the Pz electrode). The goal of the classification task is to discriminate the different patterns in the EEG. In this paradigm it is possible to infer the true label based on the reaction of the subject.

Figure 3 shows the experimental setup using this paradigm. Subjects see two different types of stimuli (target and standard) via a head-mounted display and perform a dual-task, that means, they play a virtualized labyrinth game and response to targets by pressing a buzzer.

For the data acquisition process, we had 5 subjects, with 2 recording sessions per subject. The recording sessions were divided into 5 parts, where each part yielded 720 samples of unimportant information and 120 samples of important information. For the evaluation, we used 1 part for training, and the 4 remaining parts for testing. This results in 200 evaluations ‡ for each choice of parameter in the comparison (see also Figure 3, Transfer 1).

‡ (5 subjects, 2 sessions, 5 parts for training, *4 remaining parts for testing)

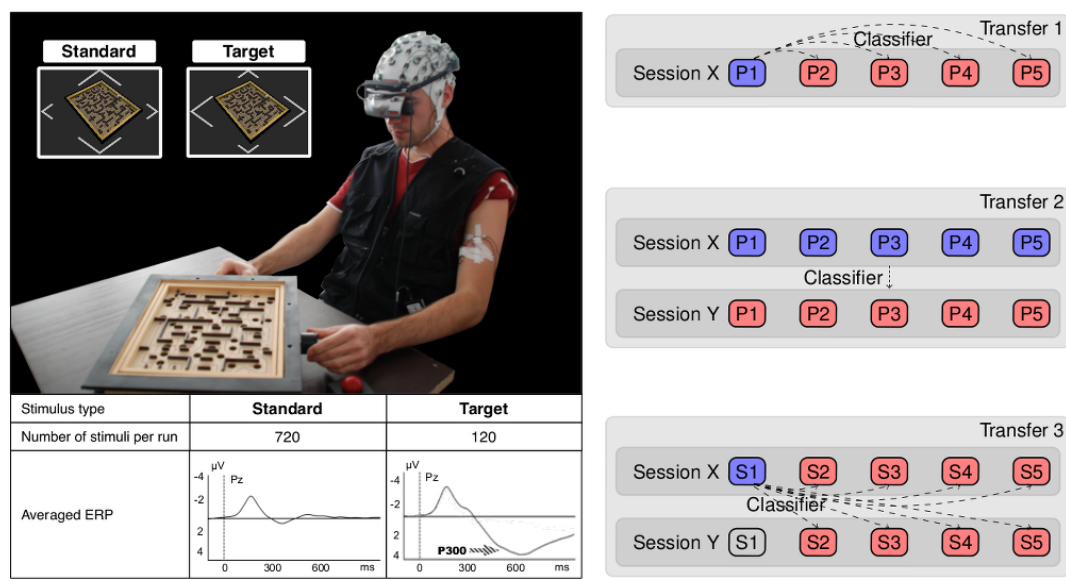

Figure 3: Setup of the data recording explained in Section 3.2 (left) and the different transfer settings (right). Pj refers to part j of the data of a recording session, whereas Sj refers to the complete recording of one session of Subject j. Blue is used for the training data and red for the testing data. The classifier is transferred from training data to the testing data as indicated by the arrows and then adapted on the testing data. The numbers 1, . . . , 5 are permuted during the evaluation in Transfer 1 and Transfer 3. For Transfer 2, they are irrelevant. Section 3 uses Transfer 1, whereas Section 4 analyzes all transfer settings.

### 3.3. Processing chain and implementation

The chosen classifier was a SVM implementation with a linear kernel, as suggested by (Hsieh et al. 2008). We limited the number of iterations to 100 times the number of samples. The regularization hyperparameter $C$ of the SVM was optimized using 5 fold cross validation, with two repetitions, and the values $[10^0, 10^{-0.5}, . . . , 10^{-4}]$.

For the synthetic data as well as for the EEG data, a feature normalization based on the training data was performed, to zero-mean and unit-variance.

In the case of the EEG data, further preprocessing was applied by means of a standard processing chain similar to the one described in (Wöhrle et al. 2015) and depicted in Figure 4 as a configuration file for the software pySPACE (Krell, Straube, Seeland, Wöhrle, Teiwes, Metzen, Kirchner & Kirchner 2013). After each stimulus presentation, a segment of one second was cut out from the data stream and normalized to zero mean and a standard deviation of one. The sampling rate of the data was reduced from 1000 Hz to 25. Then the data was low-pass filtered with a cut-off frequency of 4. The spatial filter (xDAWN) was trained on the complete training data and then applied. Afterwards, local straight lines were fitted and the respective slopes were taken as features.

For the evaluation part, each sample was first classified, and then the result was forwarded to the performance calculation routine. Afterwards, the correct label was provided to the

```
# supply node chain with data
- node: TimeSeriesSource
# standardize each sensor: mean 0, variance 1
- node: Standardization
# reduce sampling frequency to 25 Hz
- node: Decimation
  parameters:
    target_frequency: 25
# filtering with fast Fourier transform
- node: FFTBandPassFilter
  parameters:
    pass_band: [0.0, 4.0]
# linear combination of sensors to get
# reduced number of (pseudo) channels (here 8)
- node: xDAWN
  parameters:
    retained_channels: 8
# fit local straight lines and extract slope
- node: LocalStraightlineFeatures
  parameters:
             segment_width: 400
             stepsize: 120
             coefficients_used: [1]
```

Figure 4: EEG data preprocessing. This YAML specification defines the preprocessing (explained in Section 3.3) by a list of nodes (including the node name and the dictionary of parameters) for the software framework pySPACE.

algorithm for manipulating the training data of the classifier, and, if necessary, the classifier was updated. In order to save time, the SVM was only updated when the data led to a change which required an update. If, for example, data outside of the margin is added and removed, no update is required.

A preceding analysis was performed on the synthetic data to analyze the unsupervised label assignment parameter by (Spüler et al. 2012). The analysis revealed that all data should be added for the classification. Hence, we did not consider this approach for the following evaluation. For comparison, we used batch sizes $[50, 100, \ldots, 1000]$ for the synthetic data and $[100, 200, \ldots, 800]$ for the EEG data. We implemented and tested the criteria "all", "misclassified", "within margin" for adding samples, "oldest", "farthest", "border points" with tour variant for removing samples and the (optional) variants to "relabel" the data, "keep only support vectors", and "data balancing". The code is publicly available at the repository of the software pySPACE (Krell, Straube, Seeland, Wöhrle, Teiwes, Metzen, Kirchner & Kirchner 2013). For the details to the methods refer to Section 2.

We did not test an unsupervised setting for the update of the classifier. In this case either semi-supervised classifiers could be used or the classified label could be assumed to be the true label. In the latter case, all aforementioned strategies could be applied and the relabeling is especially important (Spüler et al. 2012).

For classical BCIs the true label is often not available, especially when using spellers for patients with locked-in syndrome (Mak et al. 2011). In contrast, with embedded brain reading (Kirchner 2014) the true label can be often inferred from the behavior of the subject. If the subject perceives and reacts to an important rare stimulus, the respective data can be labeled as P300 data. For simplicity, the reaction in our experiment was a buzzer press. Another example is movement prediction where the true label can be inferred by other sensors like EMG or force sensors. If possible, online learning applications should always integrate mechanisms to determine the true label in hindsight to obtain better performance.

### 3.4. Performance measure for imbalanced data

To account for class imbalance in the performance evaluation, we used *balanced accuracy (BA)* as the performance measure (Straube & Krell 2014). It is the arithmetic mean of true positive rate (TPR), and true negative rate (TNR):

$$\text{BA} = \frac{1}{2}(\text{TPR} + \text{TNR}) = \frac{\text{TP}}{\text{P}} + \frac{\text{TN}}{\text{N}} \tag{4}$$

with the number of positives (P), true positives (TP), negatives (N), and true negatives (TN). Note, that with a class ratio around $1:5$ an accuracy of 80% could be achieved by just classifying every sample as the overrepresented class. With the BA this strategy would result in a value of 50% which is equivalent to guessing. As a baseline, we used the online learning PA and the static SVM which both require to store only the linear classification vector $w$ and the offset $b$.

### 3.5. Statistical analysis

For the synthetic data, the results were analyzed by repeated measure ANOVA with 5 within-subjects factors (criteria): inclusion (ADD), exclusion (REM), data balancing (BAL), support vector handling (KSV), and relabeling (REL). For EEG data, the basket *size* was considered as an additional factor. Here, we looked for more general good performing algorithms independently from subject, session or repetition. Where necessary, the Greenhouse-Geisser correction was applied. For multiple comparisons, the Bonferroni correction was used.

In Table 1, the different strategies of how to

- **ADD**: all (a), within margin (w), misclassified (m) and
- **REM**ove: oldest (o), farthest (f), not a border point (n) samples,
- for data **BAL**ancing: no handling (n), balancing the ratio (b), or keep it fixed (k),
- **K**eeping only **S**upport **V**ectors (KSV): active (t) and not active (f), and
- **REL**abeling: active (t) and not active (f)

are compared. For each dataset, first the best combination of strategies with the respective batch size and performance (BA in percent) is given based on the descriptive analysis. The performance for the static SVM and the PA are provided as baselines. In addition, the best strategy was selected separately for each criterion which was statistically estimated irrespective of the interaction between criteria (second row for each dataset, $p < 0.05$). In case, that strategies of criteria did not differ from each other, it is not reported ["-", $p = n.s.$]. Although the best approach is different depending on the dataset, some general findings could be extracted from descriptive and inference statistics.

Table 1: Comparison of different data selection strategies for each dataset. For further details refer to the description in Section 3.5.

| DATASET | ADD | REM | BAL | KSV | REL | SIZE | PERF | SVM/PA |
|---------|-----|-----|-----|-----|-----|------|------|--------|
| Parallel | w | n | n | f | t | 150 | 96.4 | 55.9 |
|          | f | b | b | f | - |     |      | 88.7 |
| LinearShift | m | o | n | f | f | 1000 | 87.1 | 50.2 |
|             | m | o | k | - | t |      |      | 94.5 |
| Opposite | m | f | n | - | - | 200 | 95.4 | 39.4 |
|          | m | o | n | f | t |     |      | 68.4 |
| Cross | m | o | n | - | - | 150 | 95.2 | 55.7 |
|       | m | o | b | - | - |     |      | 67.6 |
| Parabola | a | o | k | f | t | 50 | 96.8 | 66.2 |
|          | m | f, o | b | t | t |    |      | 61.3 |
| 3D | a, w | o | n | - | - | 50 | 87.7 | 81.1 |
|    | w | o | n, b | f | - |    |      | 51.0 |
| EEG | m | f | b | f | f | 600 | 84.7 ± 0.3 | 83.7 ± 0.3 |
|     | w | o | b | f | f | 300, 400 | 84 ± 0.4<br>84.1 ± 0.3 | 83.0 ± 0.4 |

## 3.6. Results and Discussion on synthetic data

The results are summarized in Table 1. First, for all datasets, the best combination of strategies with a limited batch size (Table 1: first row for each dataset) outperformed the static SVM and the PA except for the "LinearShift" dataset. The SVM is static and cannot adapt to the drift and the PA is adapting to the drift but it does not forget its model modifications from previous examples.

Second, specific approaches were superior compared to other approaches (Table 1: second row for each dataset). For most cases, it was best to *add* only misclassified samples. A positive side effect of adding only misclassified samples is the reduced processing time, due to a lower number of required updates. *Removing* the oldest samples often gave good results, because a continuous shifts of a dataset leads to a shift of the optimal linear separation function and older samples would violate the separability. Not removing the border points showed bad performance in every case. Mostly, *balancing* the data led to higher performance. Often, the remaining two criteria had less influence but there is a tendency for *relabeling* data and not *keeping only support vectors*.

Third, the investigation of the interaction between the criteria could help to understand why there is some discrepancy between the best combination of strategies (first row for each dataset) and the best choice for each individual criterion (second row for each dataset). There is an interaction between some of the different criteria. For example, if a good joint selection of inclusion and exclusion criteria is made, there are several cases where relabeling is no longer necessary (possibly because the data is already well separated), or the data should not have to be balanced anymore (maybe because a good representative choice is already made). A more detailed example is given in the following, and in the discussion on EEG data.

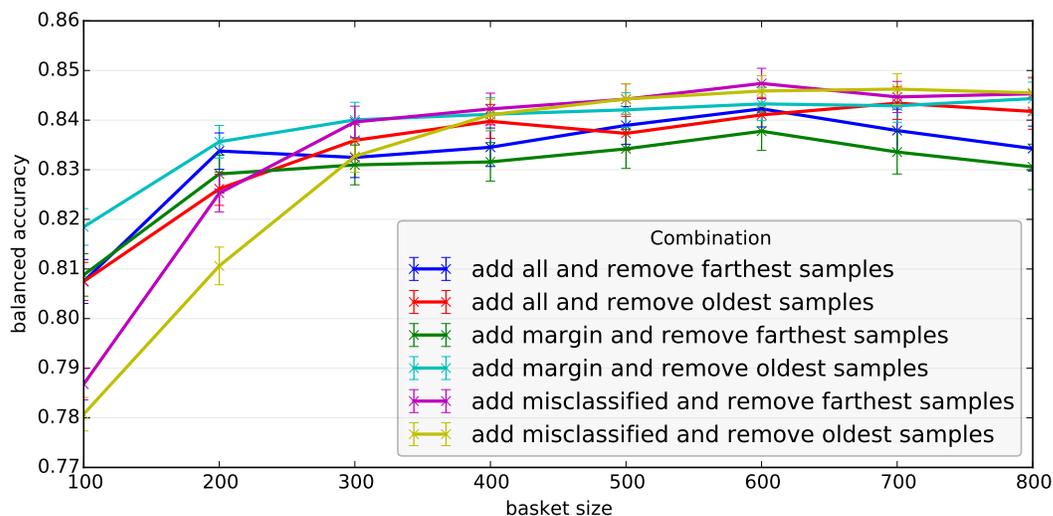

Figure 5: Comparison of all combinations of strategies from three different criteria (inclusion/exclusion/basket size) with data balancing (BAL-b) but without relabeling (REL-f) or keeping more than just support vectors (KSV-f). The mean of classification performance and standard error are depicted. For further details refer to Section 3.7.

For the "Opposite" dataset, the BAL criterion had a clear effect. All combinations with not handling the class ratios (BAL-n) were better than the other combinations with BAL-b and BAL-k. In contrast, the KSV criterion had no strong effect. In combination with KSV-t, removing no boarder points (REM-n) improved the performance but the performance decreased for a few other combinations of in-/exclusion criteria with KSV-t. On the other hand, the KSV criterion had an interaction with the relabeling criterion (REL). In combination with KSV-f, REL had no effect but REL-t improved performance for some combinations with KSV-t.

### 3.7. Results and Discussion on EEG data

For the EEG data, we observed a strong interaction between the different criteria [$p < 0.05$]. The results are summarized in Table 1. The handling of support vector affected classification performance, i.e., the performance was significantly worse when keeping only support vectors (KSV-t) compared to keeping more than just the support vectors (KSV-f). The effect of the support vector handling dominated the other criteria. The performance of combinations with KSV-f was always higher than the combinations with KSV-t. Thus, it makes less sense to choose the combinations with KSV-t. A similar pattern was observed in the handling of relabeling (REL) and data balancing (BAL). When not relabeling the samples (REL-f) and when keeping a balanced class ratio (BAL-b) the performance was superior compared to other strategies of REL and BAL. Hence, we chose a fixed strategy of these three criteria (KSV-f, REL-f, BAL-b) for the following visualization. Figure 5 illustrates the comparison of the respective combinations of criteria. The best combination was obtained when combining the inclusion of misclassified samples with the exclusion of oldest or farthest samples for the basket size of 600. This inclusion has a low number of updates. Computation for the removal of oldest samples requires very few resources whereas removing the farthest samples requires some additional effort. Hence, these combinations are beneficial in case of using a mobile device, which requires a trade off between efficiency and performance.

## 4. EVALUATION IN DIFFERENT TRANSFER SETTINGS

This evaluation is an extension of the previous comparison, focussing on different transfer scenarios in EEG data processing. First, we describe our changes in the previous evaluation setup and the respective processing chain, which is used for three different transfer setups. Then we provide the results for the different transfer settings from the P300 paradigm. The last transfer setting uses different data which was collected from a movement paradigm which elicits the lateralized readiness potential (LRP). Hence, we describe the data and the processing additionally to the results. For a better overview, we summarize all results (P300/LRP) in Section 4.6 in Table 2, and finally discuss all results together.

### 4.1. Modification of the original evaluation setup

So far, the EEG data processing only used data from a nearby session. Furthermore, the xDAWN algorithm was trained on the complete data which contradicts our original approach to only use a small batch of data. Note that the xDAWN algorithms improves with the number of samples even though it is possible to reduce this effect a bit by online learning (Wöhrle et al. 2015) or regularization (Krell, Wöhrle & Seeland 2015). The feature normalization has only a small impact on the classification performance in EEG data processing, because the preprocessing already results in a good data normalization.

Another problem of the xDAWN is its lack of transferability to EEG recordings from other sessions or subjects. For this transfer it is sometimes better to omit the spatial filter. Consequently, we do not use the xDAWN filter in the following evaluation but use the amplitudes at the different time points at all the channels directly as feature vector input. In contrast or additionally to the balancing strategies, it is also possible to set weights for the different classes in the SVM classification to balance their influence in the internal error minimization of the SVM ($t_i$). Hence, we used an SVM weight of 5 for the "Target" class (from previous experience).

In the following, we do not consider the "not a border point (n)" strategy for sample removal because its performance was too bad in the previous comparison and we do not report results on the *relabeling* strategy because it resulted in performance around guessing. Last but not least, we reduced the grid for the hyperparameter optimization due to the lack of computational resources ($[10^0, 10^{-1}, \ldots, 10^{-4}]$). This reduction is not expected to have a relevant influence to the evaluation.

The same design of statistical evaluation was performed for all transfer settings. The results from each transfer setting were analyzed by repeated measure ANOVA with five within- subjects factors (criteria): inclusion (ADD), exclusion (REM), data balancing (BAL), support vector handling (KSV), and basket size (SIZE). Where necessary, the Greenhouse-Geisser correction was applied. For multiple comparisons, the Bonferroni correction was used.

### 4.2. Results of transfer between recording parts

For this setting, we repeated the previous EEG evaluation scenario but with the already explained modified processing chain. For each choice of parameters, again we had 200 evaluations (see also Figure 3, Transfer 1). The configuration to run this evaluation is depicted in Figure 6.

For this analysis, We found no interaction of all different criteria [$p = n.s$] and can treat the criteria separately. The results showed again that keeping only the support vectors (KSF-t) resulted in significantly worse results [$p < 0.05$]. For the balancing criteria, keeping the ratio fixed was slightly better than the other criteria [$p < 0.05$]. There was no difference between balancing the data or not handling class imbalance. For the case of balancing the ratio (BAL-b) and not keeping the support vectors (KSF-f), the algorithm comparison is depicted in Figure 7. It can be seen, that in contrast to the previous analysis adding only the misclassified samples is not a good strategy anymore [$p < 0.05$ in comparison with other adding strategies] whereas the strategies of adding all or only samples within the margin show nearly similar performance [$p = n.s.$]. Even though the difference is smaller, removing the oldest samples is better than removing the farthest samples [$p < 0.05$]. Using a basket size of 600 samples performs best but it is not significantly better tan using 700 or 800 samples. These findings are also supported by the statistics over all parameter combinations.

From the descriptive analysis, there are several "best" configurations. Some are listed in Table 2 together with the performance of a static SVM and the online learning PA, which both perform worse. Furthermore, PA performs worse than the static SVM.

### 4.3. Results of transfer between recording sessions

For this transfer scenario, we concatenated the data from the five parts of a recording session to one single dataset. Then, we trained on one recoding session and performed the testing with online learning on the data from a different recording session of the same subject. This results in 10 (5 subjects, 2 sessions) evaluations for each parameter comparison (see also Figure 3, Transfer 2).

The results for KSV remained the same as in the transfer between recording parts. KSV-f is the best choice [$p < 0.05$]. For balancing (BAL) all methods showed similar performance. For the case of balancing the ratio (BAL-b) and not keeping the support vectors (KSF-f), the algorithm comparison is depicted in Figure 8.

```
type: node_chain  # Chosen subprogram (operation) in pySPACE
input_path: Transfer1_data  # relative path to the train/ test data
# each parameter is combined with the other
# and inserted in the template
parameter_ranges:
  __include__: [ADD_ALL, ONLY_MISSCLASSIFIED, ONLY_WITHIN_MARGIN]
  __basket_size__: eval(range(100,801,100))
  __exclude__: [REMOVE_OLDEST, REMOVE_FARTHEST]
  __KSV__: [False, True]
  __BAL__: [DONT_HANDLE_RATIO, KEEP_RATIO_AS_IT_IS, BALANCED_RATIO]
nodes:  # node chain template
  # previous loading preprocessing
  - node: TimeSeriesSource
  - node: Standardization
  - node: Decimation
    parameters: {target_frequency: 25}
  - node: FFTBandPassFilter
    parameters: {pass_band: [0.0, 4.0]}
  - node: TimeDomainFeatures  # simplified feature generation
  - node: GaussianFeatureNormalization
  # hyperparameter optimization
  - node: GridSearch
    parameters:
      optimization:
        ranges:
          ~~log_complexity~~: [0, -1, -2, -3, -4]
      validation_set:
        splits: 5
      evaluation:
        metric: Balanced_accuracy
        performance_sink_node:
          node: ClassificationPerformanceSink
          parameters: {ir_class: Target}
        std_weight: 0
      validation_parameter_settings: {__no_val__: false}
      final_training_parameter_settings: {__no_val__: true}
      nodes:
        - node: SorSvm
          parameters:
            add_type: __include__
            basket_size: __basket_size__
            class_labels: [Standard, Target]
            complexity: eval(10**~~log_complexity~~)
            discard_type: __exclude__
            keep_only_sv: __KSV__
            kernel_type: LINEAR
            max_iterations: 10
            retrain: __no_val__
            tolerance: eval(min(0.01*10**~~log_complexity~~, 0.01))
            training_set_ratio: __BAL__
            weight: [1.0, 5.0]
        - node: ThresholdOptimization
          parameters:
            class_labels: [Standard, Target]
            metric: Balanced_accuracy
  - node: Classification_Performance_Sink
    parameters: {ir_class: Target}
```

Figure 6: Configuration file for pySPACE in YAML for the complete comparison with the transfer scenario in Section 4.2.

Again adding only the misclassified samples is performing significantly worse than the other inclusion criteria [$p < 0.05$]. Similar to the previous transfer setting, removing the oldest samples performs significantly better than removing the farthest samples [$p < 0.05$]. When removing the oldest samples, there is no difference in performance between adding all samples and adding only samples within the margin, where the latter requires less processing resources.

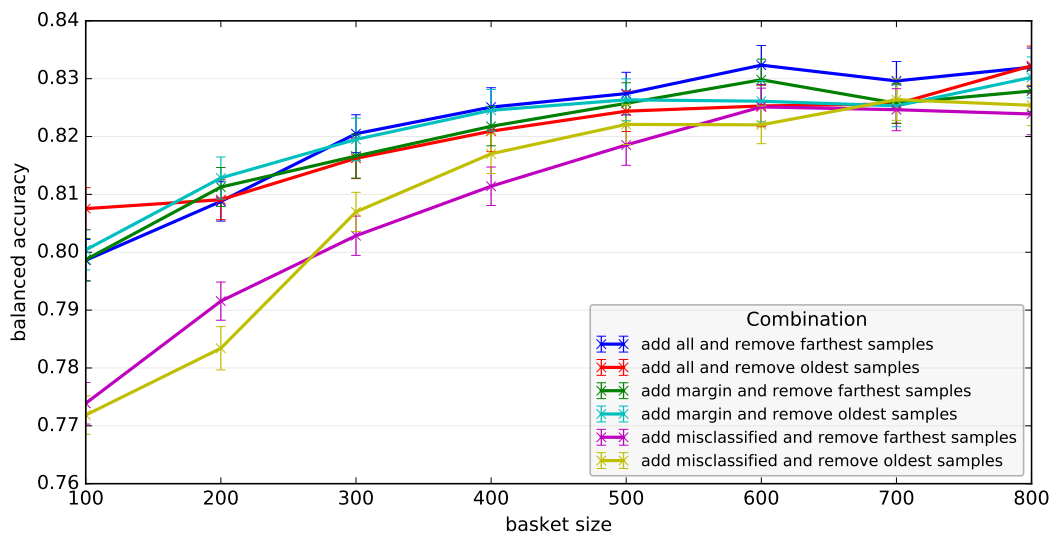

Figure 7: Transfer between *different recording parts* of the same recording session. Comparison of all combinations of strategies from three different criteria (inclusion/exclusion/basket size) with data balancing (BAL-b) or keeping more than just support vectors (KSV-f). The mean of classification performance and standard error are depicted (see also Section 4.2).

Compared to the previous transfer setting, we obtained an increase of the best (average) performance from 0.83 to 0.87 BA. Even just with a basket size of 100 samples, an (average) performance around 0.85 BA could be achieved.

From the descriptive analysis, there are several "best" configurations. Some are listed in Table 2 together with the performance of a static SVM and the online learning PA, which both perform worse. Furthermore, PA performs worse than the static SVM.

### 4.4. Results of transfer between subjects

For this transfer scenario, (as before) we concatenated the data from the 5 parts to one dataset for each recording session. Then, we trained on one recoding session and performed the testing with online learning on the data from a recording session of a different subject. This results in 80 (5 subjects, 2 sessions, 4 remaining subjects with 2 sessions each) evaluations for each parameter comparison (see also Figure 3, Transfer 3). For computational reasons, we skipped the batch sizes between 400 and 800.

The support vector handling (KSV) and the balancing (BAL) had no influence on the classification performance [$p = n.s.$]. For the case of balancing the ratio (BAL-b) and not keeping the support vectors (KSF-f), the algorithm comparison is depicted in Figure 9.

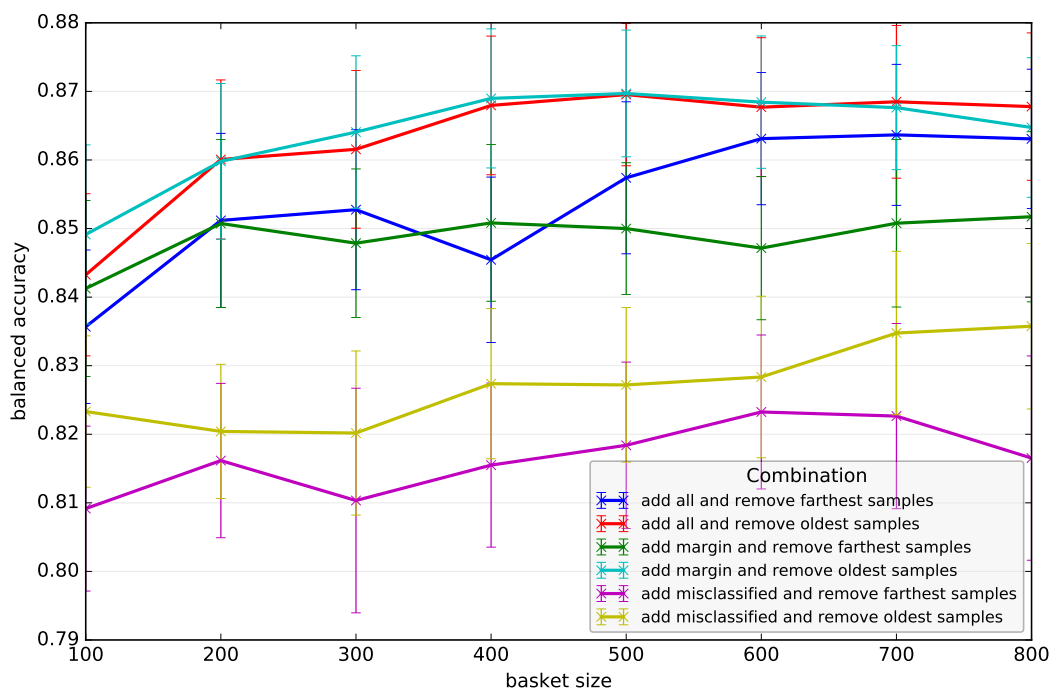

Figure 8: Transfer between *different recording sessions* of the same subject. Comparison of all combinations of strategies from three different criteria (inclusion/exclusion/basket size) with data balancing (BAL-b) or keeping more than just support vectors (KSV-f). The mean of classification performance and standard error are depicted (see also Section 4.3).

Removing oldest samples again outperforms removing the farthest samples [$p < 0.05$]. This time adding all samples performs significantly better than adding only samples within the margin [$p < 0.05$], whereas adding only misclassified samples performs again worse [$p < 0.05$].

From the descriptive analysis, there are several "best" configurations. Some are listed in Table 2 together with the performance of a static SVM and the online learning PA, which both perform worse. Furthermore, SVM performs worse than PA.

### 4.5. Transfer between different movement speeds for movement prediction

In this section, we do not analyze the EEG data from the P300 scenario but data from voluntary movements to predict an upcoming movement and to distinguish it from data were no movement occurred afterwards. This dataset is another example where the true label can be verified after the classification because if the subject is really moving this can be for example detected with EMG (Krell, Tabie, Wöhrle & Kirchner 2013), with a motion capture system (Qualisys), or the developed board. We first describe the data, the processing and transfer setup, and then provide the comparison of the SVM data selection strategies.

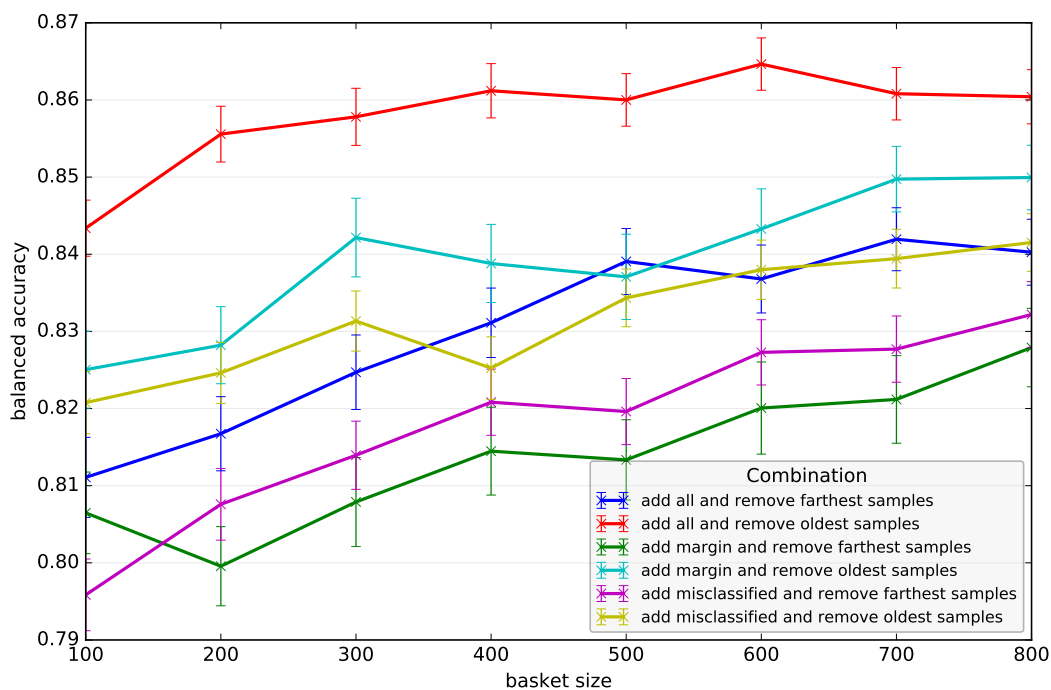

Figure 9: Transfer between sessions of different subjects. Comparison of all combinations of strategies from three different criteria (inclusion/exclusion/basket size) with data balancing (BAL-b) or keeping more than just support vectors (KSV-f). The mean of classification performance and standard error are depicted (see also Section 4.4).

*4.5.1. Data*   We used only the part of the data, which contains voluntary movements of the right forearm. Further details and information about the complete data recording setup is provided by (Tabie & Kirchner 2013, Kirchner 2014).

Eight healthy, right-handed male subjects had to place their hand on a developed board that determines the beginning and the end of the movement. For each recording run the subjects had to perform 40 voluntary movements with a break of at least five seconds between the movements to obtain a valid try. Furthermore, the subjects were first instructed to make the movements in their normal speed. Based on this, speed restrictions for fast and slow movements were calculated and afterwards recorded. For each movement speed, three consecutive runs were recorded.

We used 124 electrodes with the standard extended $10 - 20$ scheme. § As mentioned before, three different movement speeds (fast, normal, slow) were measured and each movement speed had three runs, which resulted in 120 movements per movement speed after we concatenated data of the runs. For our evaluation, we want to analyze the classifier transfer between different movement speeds. So we trained on the data from each movement speed of each subject and tested on the datasets from the remaining two speeds.

§ I1, OI1h, OI2h, I2 were used for EOG recording instead.

*4.5.2. Processing* For a simplified classification, we cut out two windows of one second length for each movement (50 ms or 200 ms before the release of the board, i.e., before the *movement onset*). For the opposite class (i.e., *no movement*), we cut out separate windows with large distance to the movement onset where the subject was not moving. This resulted in around four windows containing *no movements* for each movement. Thus, the datasets usually had around 700 samples for each movement speed, depending on the behavior of the subject.

The preprocessing was similar to the one depicted in Figure 4 but the decimation frequency was 20 Hz and as a last step in the preprocessing only the last four time points of each window were retained only the last four samples of each window were retained and directly used as features. The rest of the processing with feature normalization, SVM classification, and threshold optimization was the same with the minor difference, that a class weight of two for the movement class had to be used to again account for the class imbalance.

*4.5.3. Results of transfer between different movement speeds for movement prediction* Even though a different type of data was processed in this transfer setting, the results were similar. The number of evaluations for each parameter combination was 48 (8 subjects, 3 different movement speeds for training, 2 remaining speeds for testing).

KSV-f is the best choice [$p < 0.05$] again and for balancing (BAL) all methods showed similar performance. For the case of balancing the ratio (BAL-b) and not keeping the support vectors (KSF-f), the algorithm comparison is depicted in Figure 10.

While adding misclassified data is again the worst approach [$p < 0.05$], adding all samples clearly outperformed the approach of adding only samples which are within the margin. Removing oldest samples is better than removing farthest samples [$p < 0.05$].

From the descriptive analysis, there are several "best" configurations. Some are listed in Table 2 together with the performance of a static SVM and the online learning PA, which both perform similar worse.

### 4.6. Summary (all transfer settings on different ERP type)

In Table 2, the different strategies of how to

- **ADD**: all (a), within margin (w), misclassified (m) and
- **REM**ove: oldest (o), farthest (f) samples,
- **BAL**ance (the data): no handling (n), balancing the ratio (b), or keep it fixed (k), and
- to **K**eep only **S**upport **V**ectors (KSV): active (t) and not active (f)

are compared. For each dataset, first the best combinations of strategies with the respective batch size and performance (BA in percent) are given based on the descriptive analysis. The performance for the static SVM and the PA are provided as baselines. In addition, the best strategy was selected separately for each criterion which was estimated based on inference statistics. Note that there was no interaction of different criteria [$p < 0.05$]. and reported separately for each dataset. If the best strategy was not significantly different from other strategies these strategies are also reported in descending order. In case, that strategies of criteria did not differ from each other, it is not reported ("-", [$p = n.s.$]).

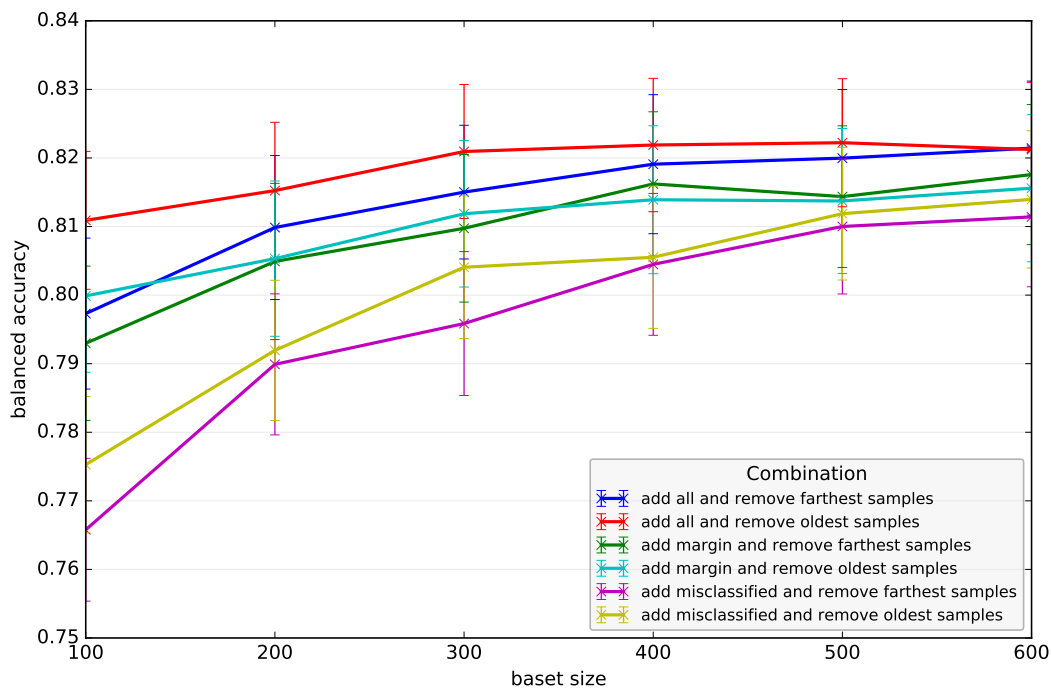

Figure 10: Transfer between *different movement speeds* for movement prediction of the same subject. Comparison of all combinations of strategies from three different criteria (inclusion/exclusion/basket size) with data balancing (BAL-b) or keeping more than just support vectors (KSV-f). The mean of classification performance and standard error are depicted (see also Section 4.5).

### 4.7. Discussion of transfer evaluation on EEG data

The methodological changes compared to Section 3 led to the changes in results in the transfer setting. For the transfer evaluation, the interaction between the different criteria was less and the results were also more clear.

The criterion of keeping only support vectors was not good because it makes the SVM to focus too much on the critical samples and forget samples which become important in the future. This is especially important for drifts which are in the normal direction of the decision hyperplane.

For all evaluations, removing the oldest data was the best exclusion approach. This behavior is reasonable because we expect drifts in the data and so more current data can model the underlying process better than the old data. Nevertheless, this observation has consequences. As already shown in the literature, it approves the necessity of online labeling and adaptation of the processing chain. But additionally we showed the necessity of real forgetting of previously learned content by the classifier. The PA does not forget but just

Table 2: Comparison of different data selection strategies for different transfer setups and their outcomes of both descriptive and inferential statistics. For further details refer to the description in Section 4.6.

| P300 | | | | | | | |
|---|---|---|---|---|---|---|---|
| Transfer | ADD | REM | BAL | KSV | SIZE | PERF | SVM/PA |
| Parts (descriptive) | a | f | b | f | 600 | $83.23 \pm 0.34$ | $82.17 \pm 0.37$ |
| | a | o | b | f | 800 | $83.22 \pm 0.34$ | $77.85 \pm 0.38$ |
| | a | f | b | f | 800 | $83.20 \pm 0.33$ | |
| (inferential) | a, w | o | k | f | 600, 700, 800 | | |
| Sessions (descriptive) | w | o | n | f | 800 | $87.39 \pm 0.88$ | $83.87 \pm 1.12$ |
| | a | o | k | f | 800 | $87.33 \pm 0.98$ | $81.59 \pm 1.21$ |
| | a | o | n | f | 800 | $87.30 \pm 0.95$ | |
| (inferential) | a, w | o | - | f | 800, 700 | | |
| Subjects (descriptive) | a | o | k | f | 800 | $86.95 \pm 0.34$ | $69.47 \pm 0.79$ |
| | a | o | n | f | 800 | $86.66 \pm 0.35$ | $78.93 \pm 0.53$ |
| | a | o | b | t | 800 | $86.47 \pm 0.38$ | |
| (inferential) | a | o | - | - | 700, 800 | | |
| LRP | | | | | | | |
| Transfer | ADD | REM | BAL | KSV | SIZE | PERF | SVM/PA |
| Speeds (descriptive) | a | o | k | f | 600 | $82.28 \pm 0.97$ | $77.36 \pm 1.31$ |
| | a | o | k | t | 600 | $82.26 \pm 0.94$ | $78.07 \pm 1.08$ |
| | a | o | n | f | 500 | $82.23 \pm 0.98$ | |
| | a | o | b | t | 600 | $82.23 \pm 0.93$ | |
| (inferential) | a | o | - | f | 600 | | |

accumulates information. So keeping a limited memory (basket) can be very helpful here.

In the previous evaluation in Section 3, adding only misclassified samples (ADD-m) was the best choice, especially because it made it possible to have only few updates. Since a sample is removed for every added sample, adding only these most important (misclassified) samples resulted in a god basket of very important samples. The reason for this is that the xDAWN spatial filter in the preprocessing was trained on the whole data and already captured most information. So the SVM could focus on the critical samples which were difficult to distinguish. But with the changed processing chain in the transfer setting, the classification task was more difficult and more data was required.

For the transfer setting, adding all (ADD-a) or adding only samples within the margin

(ADD-w) performed similar but significantly better than ADD-m when there was no big change in the data. In this case, ADD-w is the best choice because it requires less computational resources than ADD-a. But if there is a more severe drift/change in the data as it is the case for the transfer between different subjects or different movement speeds where all samples are required for the online SVM to learn the change. Probably, it is possible to use a hybrid technique, where ADD-a is used at the beginning and to switch to ADD-w, when the basket is filled with samples from the new recording session.

Even though balancing was not important in Transfer 2 and Transfer 3 (Figure 3), one should use a balancing criterion to ensure that the algorithm still works in a real applications. Since removing the oldest data is the best choice, the classifier will fail as soon as too many samples of one class arrive in a sequence. For the given data this was never the case and the class ratio remained the same.

Originally balancing the data (BAL-b) was the best choice but this time, keeping the ratio fixed (BAL-k) was better. The main reason for this difference in the results is the change in the weights of the SVM. It is an internal balancing. By using the additional balancing criterion (BAL-b) the underrepresented class would get too much weight. This indicates that the balancing needs further investigation in the future. A new criterion could be introduced to select a fixed class ratio and jointly optimize the classifier weight and the chosen class ratio. However, this might require an additional hyperparameter optimization during runtime which would consume to many resources.

The increased performance of the transfer between recording sessions and the transfer between recording parts can be easily explained with the increased amount of data (five times).

## 5. CONCLUSION

In this paper, we reviewed numerous data selection strategies and compared them on synthetic and EEG data from two different paradigms (P300 and movement prediction). As expected, we could verify that online learning can improve the performance. This even holds when limiting the amount of used data. In fact some selection criteria can even largely improve performance by forgetting too old data which disturbs the model.

Depending on the kind of data (shift), different methods are superior. Considering that usually more data is expected to improve performance or at least not to reduce it is surprising that adding only misclassified data is a good approach on our synthetic data and it is often significantly better than adding all incoming data. Furthermore, this approach comes with the advantage of requiring the lowest processing costs. If a transfer learning setting is used with bigger shifts and real world data, adding only misclassified samples is not a good approach anymore but adding all samples is appropriate, especially if the change is too big. In this case it is necessary to take all samples into consideration. Here, a hybrid approach, which detects the magnitude of the change and adapts the required inclusion criterion, is of large interest and can save resources.

Since larger drifts/transfers require to remove the oldest samples, this is in general a good choice, whereas removing only the samples which are farthest from the decision boundary does not work for larger drifts/transfers.

Our suggested variant of balancing the class ratios was beneficial in several cases, but the benefit was dependent on the type of the shift and the chosen inclusion/exclusion criterion, because some variants balance the data intrinsically, or keep the current class ratio. Hence, it should always be considered, especially since it makes the algorithms robust against long time occurrence of only one class. If the classifier was adapted to consider the imbalance of the classes for example by additional weights the class ratio should be either kept fixed or the weight should be adapted. Otherwise, a decrease in performance might occur.

To summarize we would like to repeat that it is always important to look at the interaction between the data selection strategies and that depending on the dataset and the preprocessing completely different strategies work best.

In future, we want to analyze several hybrid approaches between the approaches themselves but also between the selection strategies from SVM and PA. Some inclusion strategies can be applied to the PA and when removing samples from the training set, their weights could be kept integrated into the linear classification vector. Last but not least, different implementation strategies for efficient updates and different strategies for unsupervised online learning could be compared. In the latter, the relabeling criterion might be more beneficial than in our evaluation.

## ACKNOWLEDGEMENTS

This work was supported by the Federal Ministry of Education and Research (BMBF, grant no. 01IM14006A). We thank Marc Tabie and our anonymous reviewers for giving useful hints to improve the paper.

*Disclosure/Conflict-of-Interest Statement*

The authors declare that the research was conducted in the absence of any commercial or financial relationships that could be construed as a potential conflict of interest.

*Ethics statement*

Both studies for generating the used EEG data have been conducted in accordance with the Declaration of Helsinki and each approved with written consent by the ethics committee of the University of Bremen. Subjects have given informed and written consent to participate.